\documentclass[runningheads]{llncs}
\usepackage{graphicx}
\usepackage{subcaption}
\usepackage{textcomp}

\newcommand\Tstrut{\rule{0pt}{3ex}}         
\newcommand\Bstrut{\rule[-2ex]{0pt}{0pt}} 
\usepackage{hhline}
\usepackage{times}
\usepackage{latexsym}
\usepackage{array}
\usepackage{multirow}
\newcolumntype{C}[1]{>{\centering\arraybackslash}m{#1}}
\begin{document}
\title{A Review of Challenges in Machine Learning based Automated Hate Speech Detection}
\titlerunning{Hate speech detection challenges}
%
\author{Abhishek Velankar\inst{1,3} 
\and Hrushikesh Patil\inst{1,3} \and
Raviraj Joshi\inst{2,3}}
\authorrunning{A. Velankar et al.}
%
\institute{Pune Institute of Computer Technology, Pune, Maharashtra \and
Indian Institute of Technology Madras, Chennai, Tamilnadu \and
L3Cube Pune}
\maketitle              
\begin{abstract}
The spread of hate speech on social media space is currently a serious issue. The undemanding access to the enormous amount of information being generated on these platforms has led people to post and react with toxic content that originates violence. Though efforts have been made toward detecting and restraining such content online, it is still challenging to identify it accurately. Deep learning based solutions have been at the forefront of identifying hateful content. However, the factors such as the context-dependent nature of hate speech, the intention of the user, undesired biases, etc. make this process overcritical. In this work, we deeply explore a wide range of challenges in automatic hate speech detection by presenting a hierarchical organization of these problems. We focus on challenges faced by machine learning or deep learning based solutions to hate speech identification. At the top level, we distinguish between data level, model level, and human level challenges. We further provide an exhaustive analysis of each level of the hierarchy with examples. This survey will help researchers to design their solutions more efficiently in the domain of hate speech detection.

\end{abstract}
\section{Introduction}

The meteoric growth of social media platforms has impacted human lives in many ways. People are becoming more inclined towards the digitized mode of communication by unreservedly posting comments and messages. This often leads to the spread of harmful and violent content that may or may not be intentional. Hate speech is a specific language that is aimed toward mistreating individuals or groups with respect to their personal traits like gender, race, ethnic origin, disability, etc. The varied nature of such online content makes it more challenging to restrain its outspread. Also, a single hateful comment may largely hinder the mental as well as physical health of the person it is targeted at. It eventually reduces the belief in freedom of speech in the online media space to interact with everyone openly \cite{Isasi2017HateSI}, \cite{article}.   Hence, it is highly important to address this issue beforehand.

The main aspect that makes the filtration of online toxic content difficult is a lack of a standardized definition for hate speech that is universally accepted \cite{Kovcs2021ChallengesOH}. Though people have a fundamental perception of what hate speech is, that does not cover the knowledge required to interpret it entirely. Moreover, social media companies such as Facebook, YouTube and Twitter have made a lot of policies and rules to control its spread on their sites \cite{Ullmann2019QuarantiningOH}. But, as the extent of data being originated is immense, it is still difficult to supervise and needs a lot of attention. Also, the percentage of hateful content online is very limited as compared to the non-hateful i.e. neutral or positive content \cite{MacAvaney2019HateSD} which encourages various kinds of biases in the hate speech detection systems as well. In this paper, we try to study and explain these issues in detail. 

\begin{figure*}[hbt!]
\centering
\includegraphics[width=12cm,height=12cm,keepaspectratio]{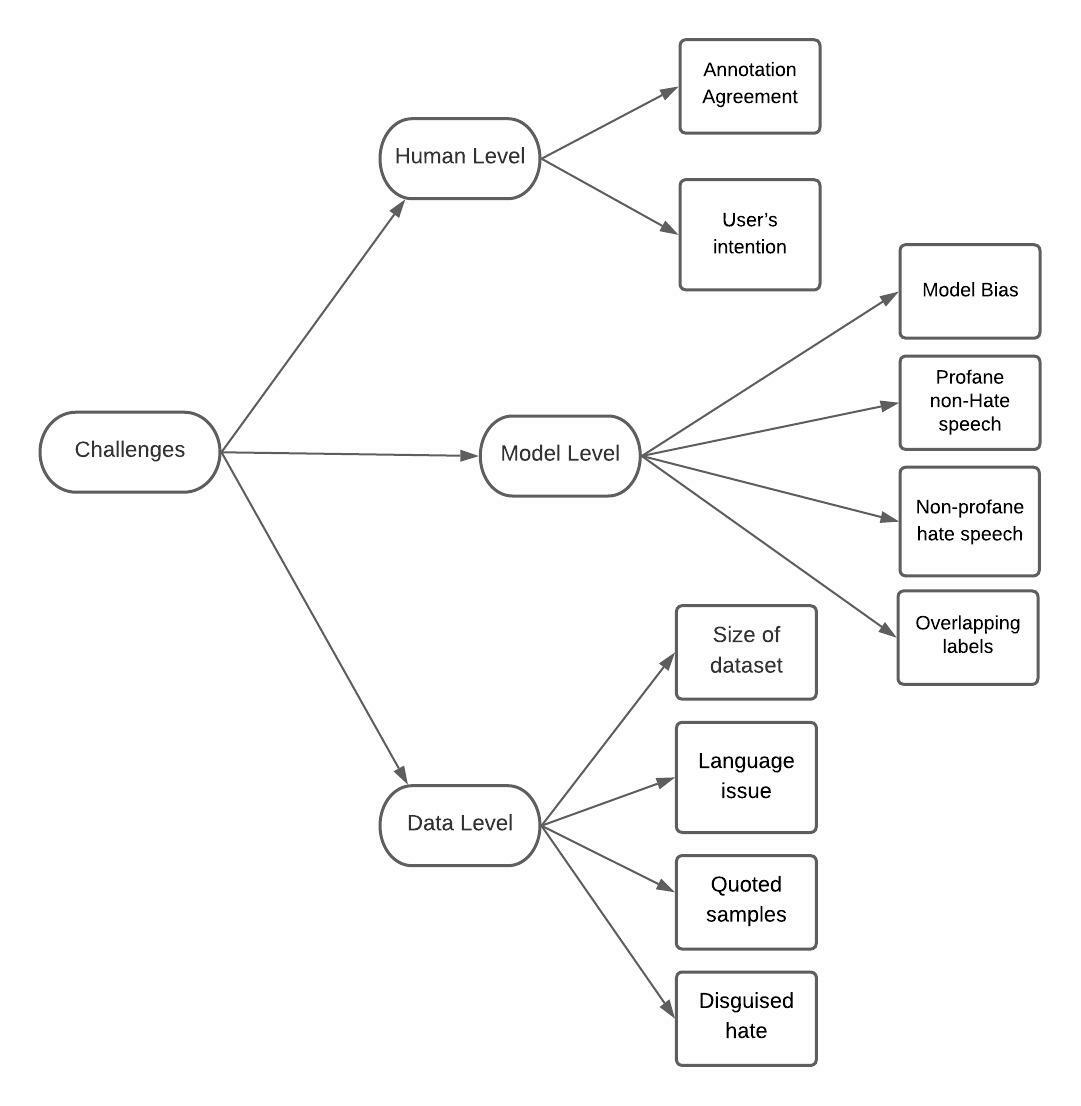}
\label{fig:1}
\caption{Classification of challenges in Hate Speech Detection}
\end{figure*}

In this work, we propose a detailed study on various challenges faced in hate speech detection on social media \cite{malmasi2017detecting,chakraborty2022nipping}. We specifically focus on challenges faced by deep learning or machine learning based solutions in the identification of hateful content \cite{badjatiya2017deep,mozafari2019bert,aluru2020deep,mandl2021overview,velankar2021hate,toraman2022large}. We divide these challenges into majorly three categories – data level, model level and human level challenges, and represent them in a hierarchical manner by providing their subtypes. Furthermore, we give a comprehensive study of how these challenges at each level of the hierarchy affect the process of filtering out hateful content by giving the case studies. This work is aimed to help the researchers to understand and study the problem of hate speech detection meticulously and give insights about the major challenges while building more efficient solutions.

\section{Related Work}

Over the last decade, the problem of hate speech has become noticeable. There has been a significant amount of work carried out to control its spread. However, the research can be seen towards addressing various challenges in detecting this content as well. 

In \cite{https://doi.org/10.48550/arxiv.2201.00961}, authors propose methodological challenges behind building automated hate speech mitigation systems. This work talks about existing challenges such as context and subjectivity of language, the multifaceted nature of hate speech, lack of sufficient data, etc. The authors discuss the possible proposed solutions for each of the challenges as well.

Bias is a major factor behind the underperformance of the system. In \cite{https://doi.org/10.48550/arxiv.2202.00126}, the authors propose a detailed survey of various unintended biases that hinder the model’s performance and create a taxonomy for the biases as well. The categorization of biases has been made based on sources of harm during the data collection process and targets of harm such as gender, racial and psychographic bias. Other works in the area of bias mitigation include \cite{blodgett-etal-2020-language}, \cite{app11073184} and \cite{https://doi.org/10.48550/arxiv.2112.04359}.

In \cite{https://doi.org/10.48550/arxiv.2109.02941}, the authors provide a conceptual framework for countering hate speech on social media. They explore the causes and consequences of the spread of hate speech and study proactive and reactive methods of hate speech mitigation as well. Lastly, they provide a detailed study of ethical and legal challenges in countering hate speeches.

\section{Challenges in hate speech detection}
We have categorized the challenges in hate speech detection broadly into three categories – data level, model level and human level. The detailed hierarchical representation of the above categories and their subtypes is shown in figure 1. Also, an example text for different subcategories is added in Table \ref{annot-ex}.

\subsection{Data level challenges}

Data is the foremost entity in any machine learning task. It plays a crucial role in designing the system that fulfills the needs of the task convincingly. The quality of data used for training is directly proportional to the ability of the model to adapt to real-world use cases. When data itself is problematic, it becomes highly challenging to build an efficient system. The detailed subcategories in this section of challenges are explained below.

\subsubsection{Size of the dataset.}

There are a limited number of datasets available publicly which provide samples in a wide range of hateful categories. Also, collecting a large number of hateful samples is a tedious task as the size of non-hateful content is ordinarily oversized \cite{https://doi.org/10.48550/arxiv.2201.00961}. More importantly, as discussed earlier, there is no universal definition for hate speech available. Thus, to agree upon certain comments are hateful is burdensome. Furthermore, getting the right amount of data for regional languages that are generally low in resources is even harder \cite{romim2021hate,velankar2022l3cube,nozza2022hate}. These issues make it more challenging to get equitable data for building efficient systems.

\subsubsection{Language issues.}

\textbf{Representation}: The data represented in the English language is more straightforward for computers to represent because of the standard ASCII codes available. When it comes to other natural languages, building a machine to effectively recognize these texts becomes an oppressive task and needs additional mechanisms to counter this issue. \newline
\textbf{Use of local languages in text}: Every language has different dialects in different regions. There is always a slight change in words and style of writing. The traits like the use of delimiters, order of words, grammatical variations, etc. change with different languages \cite{Harish2020ACS}.  This problem becomes even worse in morphologically rich languages as regional dialects change by large amounts, making the dataset even more complex. \newline
\textbf{Use of foreign language}: In today’s social media world, the use of code-mixed language while texting is a common practice. To reduce the extent of code-mixed data, text written directly in foreign alphabets could be removed while preprocessing. But it can clutter an entire sentence thereby disturbing its context and meaning.

\subsubsection{Quoted samples.}

In some cases, text samples are quoted by a news handle or a third person to react to someone else’s original post or to take others' opinions, ex. retweeting on Twitter. For example, \textit{``Muslims should be sent to their Jannat", words from MLA}, these posts are generally from news handles. They have no intention to spread hate and can be considered non-hateful. But, there is a high chance that the system may filter them as hateful by looking at the original text, thereby increasing a false alarm and reducing the recall of the system. 

\subsubsection{Disguised hate speech.}

As the automated detection of hate speech is bringing about a lot of prevalence, people using social media are now to some extent aware of how this toxic content is filtered out. The easiest approach is the bag-of-words used in this case. Now, any user posting hate comments and aware of the filtration process, tries to disguise the bad word he is using by altering some characters and replacing them with a symbol or omitting or interchanging the vowels \cite{10.1145/3319535.3363271}, \cite{https://doi.org/10.48550/arxiv.1808.09115}. The most common technique is to replace characters with a star (*), a dot (.) or an underscore (\_), like \textit{Don’t teach me, go f**k yourself} or using \textit{fcuk}. This makes the detection challenging and conveys the need to build more concrete systems. It even gets perplexing when hateful text is posted as an image making it impossible to detect by normal NLP techniques. This problem can be solved by OCR but this multi-modal discussion is beyond the scope of this work.

\subsection{Model level challenges}

Machine learning models are the key factor in the entire process of hate speech detection. The better the performance of the model, the more promising our solution is towards countering hate speech. However, due to some minuscule issues in the data that is being used for training, models may tend to behave differently thereby giving undesirable results. Following are some common issues faced in building these models– 

\subsubsection{Biased model.}

A model is considered biased when it is more error-prone towards a label or group of labels. This phenomenon mainly occurs when there is a presence of an undersampled dataset. At times, there is not enough representation of a particular class provided in the dataset, resulting in bias, i.e. in \cite{10.3389/fdata.2020.00003}, a more toxic score has been given to sentences containing female/women references than male/men. This is because of the under-representation of samples which are more toxic toward male/men. Another issue is that, NLP models trained on the huge amount of data may contain stereotype information, such as \textit{He is a doctor} or \textit{She is a nurse} this makes the model think that \textit{doctor} is related to \textit{He} and \textit{nurse} is related to \textit{She}, even though they are completely unrelated. This problem can be resolved by removing samples representing these type of relationships. Also, the embedding representations can be modified to remove this kind of information.

\subsubsection{Non-profanic hate speech.}

Sometimes samples contain sentences which are inclined towards hate or profanity, but it doesn’t contain any bad or profanic words. These samples mainly target communities with respect to their race, gender, ethnic origin, etc. For example, \textit{Black people should not be allowed to live in this country}.  As the most elementary way to detect hate speech is the bag-of-words approach where, if a sentence contains words from a particular bag, it would be labeled accordingly. In this case, there are high chances that these types of text will go undetected and will degrade the performance of our system.

\subsubsection{Profane non-hate speech.}

Sometimes a user posts a comment containing profane words, but the user has no intention to abuse or hurt anyone. The user may be referring to the situation or joke or sarcasm. The comment generally comes out of excitement. As the profane words are used, the non-harmful comment will also be filtered as hate by the model. These types of samples may trigger false alarms. For example, \textit{He is f**king idiot} vs. \textit{He is f**king genius}. Here, we can see how the interpretation of words changes drastically with the sentence. These type of comments are generally associated with sports events. 

\subsubsection{Overlapping labels.}

In most cases, there arises a problem that the sample that has to be labeled, exhibits properties of two or more classes, this is termed as class overlap \cite{6753930}. The minor differences between the samples are not enough to capture the characteristics of the sample by following data annotation policies. Hence annotators have to randomly assign any of the possible classes as there is approximately an equal probability for the sample to be in any of the classes. This creates a problem for the model while training, to memorize and recognize patterns in given inputs.

\subsection{Human Level challenges}

Humans are considered as the most intelligent species on earth and are good at pinpointing the appropriate hypothesis. But, when it comes to the problem of hate speech detection and its diversified nature, at times it becomes tricky to draw any conclusions as well. In this section, we try to address these issues while dealing with this domain of hate speech.

\subsubsection{Annotation agreements.}

As there is a lack of a universal definition for hate speech, different people interpret the magnitude of hate in their own way. Hence, a particular sentence can be toxic to a set of people which might not be the case for others. This becomes a compelling issue while manually labeling the dataset. Annotators from different regions with different backgrounds may interpret a particular kind of hate speech in contrasting ways. This results in disagreements in the annotation process giving a low kappa score for the dataset. Furthermore, the accuracy of the system that is trained on a particular dataset is largely dependent on the annotation of the dataset which makes it imperative to counter this issue. 

\subsubsection{User intention.}

One of the toughest tasks in the annotation process is to comprehend the intention or context of the user behind any harmful comment. As there is a huge difference between one, in reality, posting virulent comments and being sarcastic, it gets puzzling to come to the right conclusion \cite{Harish2020ACS}, \cite{Kovcs2021ChallengesOH}.  This requires a lot of understanding of the current situation, political background if any and history of the post. This also obligates one to check facts regarding the post before labeling it. E.g. \textit{Hitler had the best solution for Jews}. This sentence has no bad or hate words, but by knowing the historical context we can understand that this is hate speech. Praising or supporting the people spreading hate speech should also be classified as hate speech.

\begin{table*}
\caption{\label{annot-ex} Example text for various categories of hate speech challenges.}
\setlength\tabcolsep{10pt}
\renewcommand{\arraystretch}{4}
\centering
\resizebox{\columnwidth}{!}{%
\begin{tabular}{|c|c|c|}
\hline
Category & Sub-category & Example\Tstrut\Bstrut\\
\hline
\multirow{2}{*}{Data level} &  Use of foreign language (code switching) & \parbox{6cm}{\Tstrut a. \textit{Kisi mahan purush ne kaha hai} (A great person has said), “War never decides who is right, it only decides who is left”. \\ b. \textit{Jago India jago...} (wake up India wake up...) See that if how badly reservation is showing its worst effect...\\ }\\  
\hhline{~--}
& Quoted samples & \parbox{6cm}{\Tstrut a. \textit{``Muslims should be sent to their Jannat"}, words from MLA. \\ b. \textit{``We will destroy you and your family"}, a threat on Facebook. \\} \\
\hline
\multirow{4}{*}{Model level} & Disguised hate speech & \parbox{6cm}{\Tstrut a. Burn this perverted f**ker!  Put him in prison!!!\\ b. If he kept winning like this we are so fcuked. \\ }\\
\hhline{~--}
& Non-profane hate speech & \parbox{6cm}{\Tstrut a. ``Black people should not be allowed to live in this country”. \\ b. People from majjid are reason behind this violence. \\} \\ 
\hhline{~--}
&  Profane non-hate speech & \parbox{6cm}{\Tstrut a. He is f**king lengend of rap songs. \\ b. No f**king way this will happen. \\} \\
\hhline{~--}
& Overlapping samples & \parbox{6cm}{\Tstrut a. India cannot process due to these raskal politicians. \\ b. Bastards are harrassing poors just to fill their pockets. \\} \\
\hline  
\multirow{1}{*}{Human level} & User intention or context & \parbox{6cm}{\Tstrut a. Hitler had the best solution for Jews. \\ b. Send you mother to me, ill teach her. \\} \\
\hline
\end{tabular}
} 
\end{table*}  

\section{Conclusion}

In this work, we have provided a detailed survey of a wide range of challenges faced in hate speech detection tasks. The research has been carried out by segregating these challenges into three broad categories, namely data level, model level and human level. These categories are further divided into more detailed subcategories and explored by giving examples for each. The study shows that the issue of the spread of hate speech and its detection is still a challenging case and needs to be addressed appropriately in order to produce worthwhile results.

\section*{Acknowledgements}
This work was done under the L3Cube Pune mentorship
program. We would like to express our gratitude towards
our mentors at L3Cube for their continuous support and
encouragement.

%
%
%
%
\bibliographystyle{splncs04}
\bibliography{main}




\end{document}